\documentclass[11pt]{article}

\usepackage[final]{acl}

\usepackage{times}
\usepackage{latexsym}

\usepackage[T1]{fontenc}
\usepackage{amsmath}
\usepackage[utf8]{inputenc}

\usepackage{microtype}

\usepackage{inconsolata}

\usepackage{graphicx}
\usepackage{subcaption}
\title{Online Density-Based Clustering for Real-Time Narrative Evolution Monitoring}

\author{
  \textbf{Ostap Vykhopen},
  \textbf{Viktoria Skorik},
  \textbf{Maksym Tereshchenko},
  \textbf{Veronika Solopova}\\
  \texttt{ostap.vykhopen@mantisanalytics.com} \\}
\begin{document}
\maketitle
\begin{abstract}
Automated narrative intelligence systems for social media monitoring face significant scalability challenges when relying on batch clustering methods to process continuous data streams. We investigate replacing offline HDBSCAN with online density-based clustering algorithms in a production narrative report generation pipeline that processes large volumes of multilingual social media data. While HDBSCAN effectively discovers hierarchical clusters and handles noise, its batch-only nature requires full retraining for each time window, limiting scalability and real-time adaptability.
We evaluate online clustering methods with respect to cluster quality, computational efficiency, memory footprint, and integration with downstream narrative extraction. Our evaluation combines standard clustering metrics, narrative-specific measures, and human validation of cluster correctness to assess both structural quality and semantic interpretability. Experiments using sliding-window simulations on historical data from the Ukrainian information space reveal trade-offs between temporal stability and narrative coherence, with DenStream achieving the strongest overall performance. These findings bridge the gap between batch-oriented clustering approaches and the streaming requirements of large-scale narrative monitoring systems.
\end{abstract}

\section{Introduction}
Automated narrative intelligence systems have become essential tools for monitoring information ecosystems, particularly in contexts requiring rapid detection of emerging narratives, misinformation campaigns, and geopolitical discourse patterns. Organizations ranging from defense agencies to public health authorities rely on these systems to process high-volume social media streams and generate actionable intelligence reports. However, the computational architecture of such systems often reflects fundamental tensions between the batch-oriented nature of modern machine learning frameworks and the inherently streaming character of social media data.
This tension becomes particularly acute when examining HDBSCAN \cite{mcinnes2017hdbscan}, a clustering algorithm widely adopted for narrative discovery, especially popularised by BerTopic library \cite{scarpino2022investigating,eklund2023empirical}.

Despite its theoretical strengths, handling variable-density clusters, noise, and hierarchical structure, HDBSCAN introduces significant operational bottlenecks in production narrative monitoring. Its strict batch processing requires loading all document embeddings into memory for each clustering run, leading to high memory usage and repeated O(NlogN) computations. In daily monitoring settings, this necessitates full re-clustering of incoming data, discarding previously learned structure even when narratives evolve gradually. In practice, maintaining stable cluster granularity further requires frequent parameter tuning as data volume fluctuates. Most critically, HDBSCAN cannot support incremental updates, preventing real-time cluster assignment and progressive narrative refinement. As historical data accumulates, these limitations render batch re-clustering increasingly impractical at scale.
These constraints motivate our central research question:
\textit{Can online clustering algorithms replace HDBSCAN in narrative monitoring pipelines while preserving cluster quality, maintaining integration compatibility, and improving computational efficiency?}

We evaluate online density-based clustering methods that automatically infer the number of clusters, remain robust to noise in social media streams, and support efficient incremental updates with substantially lower memory and computational overhead than batch re-clustering. Our contributions are threefold:
\begin{enumerate}
\item We propose a realistic evaluation framework for streaming clustering, incorporating temporal pretraining, incremental daily updates, and sliding-window simulations on large-scale multilingual social media data.
\item We systematically compare online density-based methods (DBSTREAM, DenStream) against a batch HDBSCAN baseline within a production-style pipeline, using a unified evaluation that combines standard clustering metrics with narrative-specific measures.
\item We analyze scalability and correctness issues in widely used reference implementations and demonstrate their impact on real-world performance. Based on empirical and operational evidence, we identify DenStream as the most viable online alternative and outline the system adaptations required for deployment in narrative monitoring pipelines.
\end{enumerate}

\section{Theoretical Background and Related work}
\subsection{Density-Based Clustering}

Density-based clustering, introduced with DBSCAN \citep{ester1996density}, identifies clusters as dense regions separated by sparse areas, enabling the discovery of arbitrarily shaped clusters while robustly handling noise. Unlike partition-based methods such as k-means or hierarchical agglomerative clustering, DBSCAN requires only two parameters—the neighborhood radius $\epsilon$ and a minimum density threshold—and does not require predefining the number of clusters.

HDBSCAN extends DBSCAN by removing the need to specify $\epsilon$ through hierarchical clustering based on mutual reachability distances and cluster stability. By extracting stable branches from a minimum spanning tree, HDBSCAN automatically determines cluster granularity, supports varying-density clusters, and provides soft cluster assignments via outlier scores.

Despite these advantages, DBSCAN and HDBSCAN are inherently batch-oriented, requiring access to the full dataset to construct distance structures and extract clusters. This assumption limits their applicability to streaming settings, where data arrives continuously and full reprocessing is impractical. In narrative monitoring applications, where topics emerge, evolve, and fade over time, clustering methods must adapt incrementally without predefined cluster counts, motivating the use of online density-based alternatives.

\subsection{Online and Streaming Clustering Algorithms}

\subsection{Streaming Clustering}

Streaming clustering addresses scenarios in which data arrives continuously and memory constraints preclude storing the full dataset. Early work such as CluStream \cite{chan1982updating,caussinus1982computing} introduced a two-phase framework combining online micro-clustering with offline macro-clustering. While CluStream enables temporal analysis of evolving streams, it requires the number of clusters to be specified in advance and is best suited to spherical cluster structures.

Subsequent density-based approaches remove this limitation. DBSTREAM \cite{hahsler2016clustering} maintains a shared density graph between micro-clusters, allowing final clusters to be reconstructed based on density connectivity rather than distance alone. DenStream \cite{cao2006density} extends this idea by incorporating temporal decay through core and outlier micro-clusters, enabling adaptation to concept drift while supporting arbitrary cluster shapes. However, this flexibility introduces additional complexity in managing micro-cluster lifecycles and decay parameters.

More recent neural clustering approaches combine representation learning with clustering objectives, but they typically require large training datasets and offer limited interpretability, which constrains their applicability in analyst-facing narrative monitoring systems.

\subsection{Streaming Topic Modeling}

Traditional topic models such as LDA and NMF have been extended to online settings through incremental inference and hybrid approaches \cite{das2025elda}. However, these methods primarily rely on bag-of-words representations and do not naturally capture contextual semantic similarity. Recent work therefore integrates topic modeling with dense embeddings to improve semantic coherence \cite{yadav2025hybrid}.

An alternative embedding-first paradigm leverages transformer-based representations for streaming topic discovery \cite{angelov2020top2vec,bianchi2021pretraining}. In this framework, documents are embedded into a dense semantic space, clustered to identify coherent topics, and subsequently labeled or summarized using language models. By decoupling representation learning from cluster assignment, this approach captures deep semantic relationships while remaining compatible with incremental clustering, making it well suited for large-scale narrative monitoring over continuously arriving data streams.

\section{Methodology \& Experimental setup}
\begin{figure*}[h]
    \centering
    \includegraphics[width=0.6\textwidth]{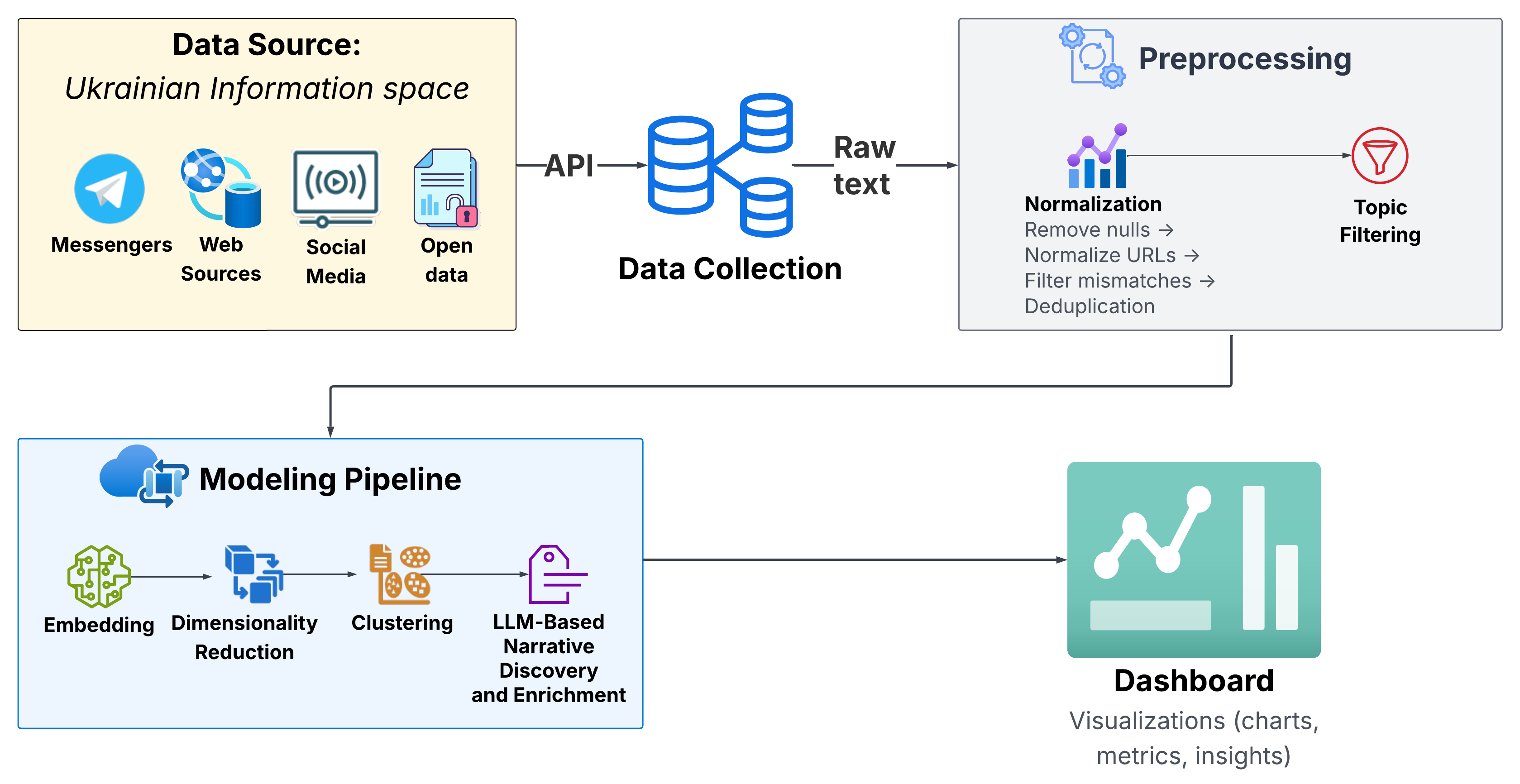}
    \caption{Four stage pipeline for topic and narrative discovery.}
    \label{fig:myfigure}
\end{figure*}
\subsection{System Architecture}
Our narrative intelligence system implements a four-stage pipeline for automated topic discovery and monitoring in social media streams.
\paragraph{Data Collection and Preprocessing.} Raw documents are collected from Ukrainian information space sources, including news portals, Telegram channels, and user comments. Text undergoes standard preprocessing including normalization, and language detection filtering.
\paragraph{Modeling Pipeline.} The core topic modeling pipeline consists of four sequential stages, and is illustrated in Figure \ref{fig:myfigure}:
\begin{enumerate}
    \item Embedding Generation: For all experiments, the documents are transformed into dense semantic representations using multilingual MiniLM\footnote{sentence-transformers/all-MiniLM-L6-v2} model from sentence transformers library.
    \item Dimensionality Reduction: Document embeddings undergo dimensionality reduction (typically via UMAP), preserving local neighborhood structure while reducing computational complexity for downstream clustering
    \item Clustering: Documents are grouped into topically coherent clusters based on their reduced embeddings. This is the component under investigation in our experiments
    \item LLM-based Topic Labeling and Enrichment: to isolate the influence of clustering algorithms, a proprietary LLM, with stable parameters and prompt was used to generates human-interpretable topic labels and summaries for each discovered cluster in all experimental settings.  
\end{enumerate}
The final component provides operational intelligence visualizations and reporting interfaces for analysts
This pipeline architecture decouples representation learning from cluster discovery, enabling modular replacement of clustering algorithms while maintaining embedding quality and downstream interpretation capabilities.
\subsection{Experiment Design}
\paragraph{Dataset} We evaluate online clustering algorithms using data from the Ukrainian information space, including news portals, Telegram channels, and user comments. The dataset\footnote{The data and code can be shared upon request to vetted academics for reproducibility purposes.} exhibits typical characteristics of narrative monitoring applications: high volume ($\approx$17,000 documents per day), topical diversity reflecting ongoing geopolitical discourse, and substantial noise from low-quality or off-topic content.
\begin{table*}[t]
\centering
\caption{Clustering Performance Comparison. The table reports clustering quality metrics for the baseline HDBSCAN algorithm and two online clustering approaches evaluated on daily data from the Ukrainian information space (approximately $N \approx 17{,}000$ documents per day before cleaning and filtering).}
\label{tab:clustering_performance}

%\small
%\setlength{\tabcolsep}{4pt}
\begin{tabular}{lccccc}
\hline
\textbf{Algorithm} & \textbf{Silhouette} & \textbf{DBI} & \textbf{Distinct.} & \textbf{Contingency} & \textbf{Variance} \\
\hline
HDBSCAN & 0.592 & 0.550 & \textbf{0.352} & \textbf{0.299} & \textbf{0.286} \\
DBSTREAM           & 0.327 & 1.220 & 0.266 & 0.630 & 0.459 \\
DenStream          & \textbf{0.685} & \textbf{0.453} & 0.319 & 0.389 & 0.319 \\
\hline
\end{tabular}
\end{table*}

\begin{table*}[t]
\centering
\small
\begin{tabular}{lcccc}
\hline
\textbf{Algorithm} & \textbf{Train (s)} & \textbf{Pred. (s)} & \textbf{Clusters} & \textbf{Narratives} \\
\hline
HDBSCAN  & 13   & 1.3 & 1063 & 118 \\
DBSTREAM & 232  & 127 & 266  & 58  \\
DenStream & 3.56 & 1.7 & 303  & 53  \\
\hline
\end{tabular}
\caption{Operational Metrics. The number of narrative candidates does not represent the actual number of narratives, but rather the number of clusters classified as narratives by the model. This distinction should be considered when interpreting the results.}
\label{tab:operational_metrics}
\end{table*}
\begin{figure*}[t]
    \centering
    \begin{subfigure}[t]{0.45\textwidth}
        \centering
        \includegraphics[width=\textwidth]{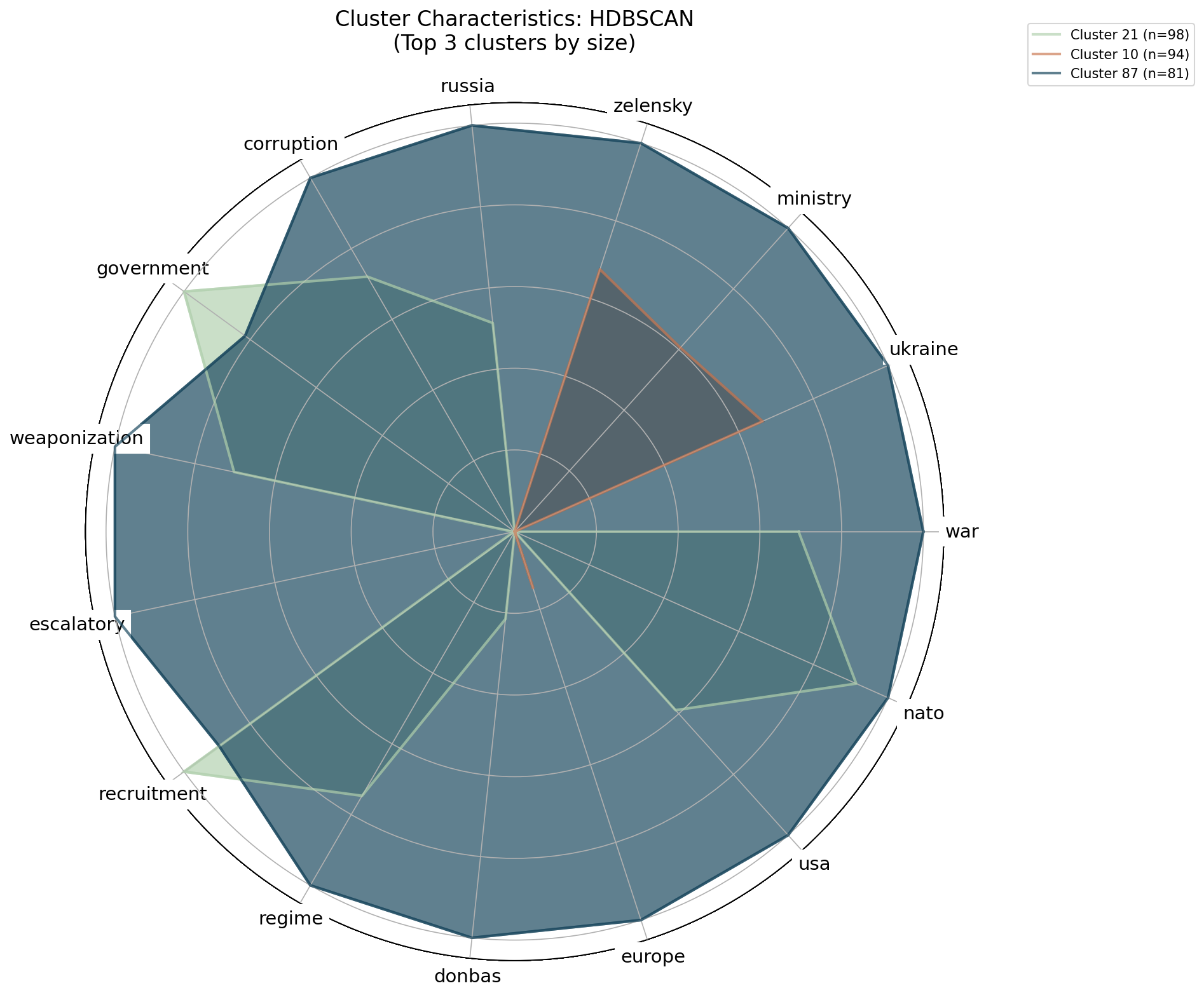}
        \caption{HDBSCAN}
        \label{fig:hdbscan}
    \end{subfigure}
    \hfill
     \begin{subfigure}[t]{0.45\textwidth}
        \centering
        \includegraphics[width=\textwidth]{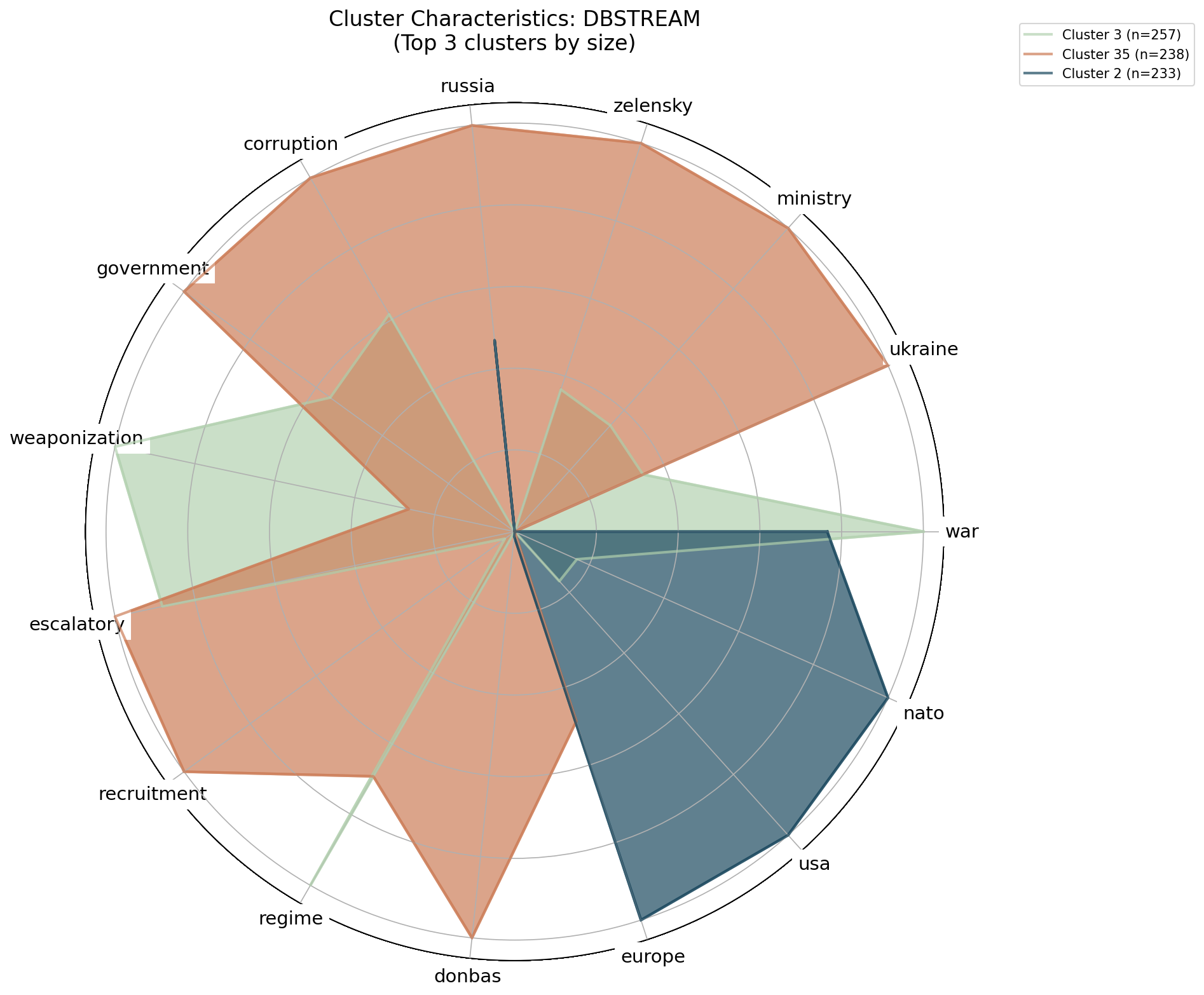}
        \caption{DBSTREAM}
        \label{fig:dbstream}
    \end{subfigure}
    \hfill
    \begin{subfigure}[t]{0.45\textwidth}
        \centering
        \includegraphics[width=\textwidth]{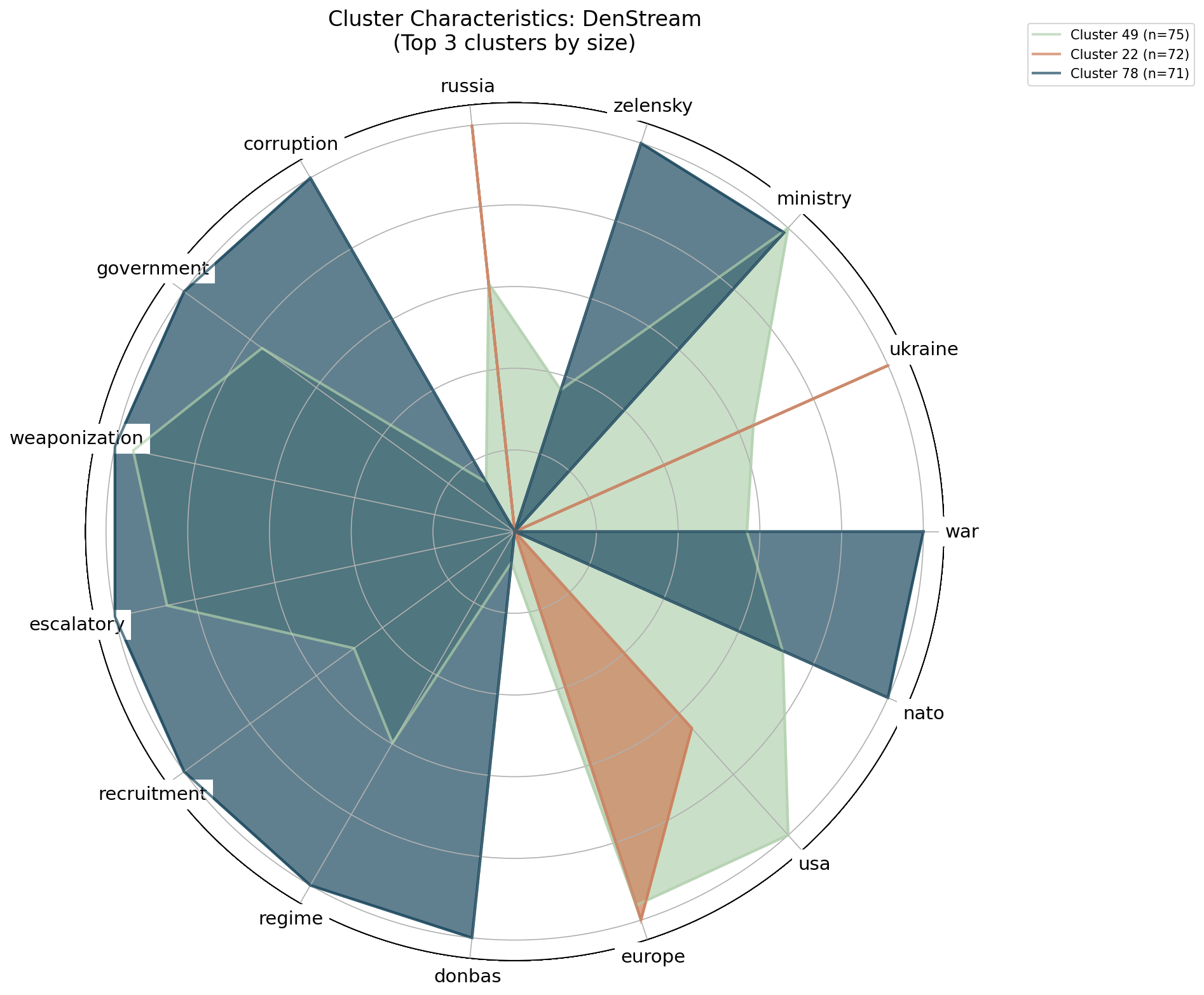}
        \caption{DenStream}
        \label{fig:denstream}
    \end{subfigure}
    \hfill
    \caption{Comparison of online clustering methods: (a) HDBSCAN, (b) DBSTREAM, and (c) DenStream. Each fingerprint represents a cluster projected onto a set of domain-specific keywords, forming a polygon that encodes its semantic profile. Sharp spikes indicate focused narratives, while broad shapes show multi-topic or hybrid narratives. Overlapping areas reveal shared or competing narratives, and may also indicate overlapping clustering.
The charts above show the Semantic Fingerprints of three clusters, based on a set of highly relevant keywords in the current Ukrainian information space. Chart (A) (HDBSCAN) shows that the first cluster covers nearly all keywords, which may indicate contamination by a mix of different narratives. In contrast, DBSTREAM and DenStream display a clearer per-keyword distribution and better separation between clusters.
}
    \label{fig:clustering_comparison}
\end{figure*}
For the experiments, data are split to reflect realistic streaming conditions. A pretraining set of 68,917 documents collected over six days (01.05–01.11) is used to initialise the models, followed by a streaming training day comprising 11,466 documents from 01.12, on which online updates and clustering behaviour are evaluated.
\paragraph{Baseline System.} Our baseline employs HDBSCAN operating in strict global mode, where clustering is performed independently on each day's data without reference to historical patterns. Specifically, for a target day d, HDBSCAN processes only the documents from day d, discovering clusters de novo without leveraging any structure learned from previous days. This approach represents the current production paradigm where each daily report requires complete re-clustering.
\paragraph{Online Algorithm Evaluation.} We evaluate three online clustering algorithms from the River library: DBSTREAM, DenStream, TextClust. Each online algorithm follows an incremental training protocol: (1) initialization on historical context (6 days of data preceding the target day), and (2) incremental updating and prediction on the day of interest. This approach simulates a realistic production deployment where the system has access to recent historical context before processing new daily data streams.

\paragraph{Evaluation Metrics.} We assess clustering quality using three standard metrics computed on the day of interest: \textbf{(i) Silhouette Score} [-1, 1], which measures cluster cohesion and separation, with higher values indicating better-defined clusters; \textbf{(ii) Davies-Bouldin Index} [0,  $\infty$], measuring average similarity between clusters and their most similar cluster, with lower values indicating better separation; \textbf{(iii) Narrative Distinctness} \cite{irani2025discourse}, which quantifies how distinct a narrative is by combining the average and minimum cosine distances between its centroid and all other narrative centroids and \textbf{(iv) Narrative Contingency and (v) narrative variance} \cite{upravitelev2025exploring}, measuring the cosine distance between a centroid and the top-ranked embedding from the set and showing the variance of text embeddings relative to their centroid, respectively.
 $\sigma^2 = \frac{1}{m_i} \sum_{j=1}^{m_i} \|\mathbf{t}_{ij} - \mathbf{c}_i\|_2^2$ 

Additionally, we measure computational efficiency through training time, prediction time, and the number of discovered clusters. These operational metrics directly address the scalability concerns motivating our investigation.
\section{Results}
We demonstrate our results in Table \ref{tab:clustering_performance} and \ref{tab:operational_metrics}, while the difference in semantic fingerprints of the top 3 clusters of each algorithm can be seen in Figure \ref{fig:clustering_comparison}.

\subsection{Cluster Quality Metrics}
DenStream demonstrates superior cluster quality across conventional clustering metrics. It achieves the highest Silhouette Score (0.685), indicating well separated and cohesive clusters, representing improvement over the HDBSCAN baseline (0.592). Similarly, DenStream attains the lowest Davies-Bouldin Index (0.453) compared to the baseline (0.550), indicating better cluster separation with lower inter-cluster similarity. 
DBSTREAM exhibits poor performance on standard clustering quality metrics. Its Silhouette Score (0.327) is notably lower than the baseline (0.592), while its Davies-Bouldin Index (1.220) is more than twice the baseline value (0.550), indicating substantial inter-cluster confusion. This suggests DBSTREAM's density based micro-cluster approach may be insufficiently granular for the high dimensional semantic space of document embeddings.
\subsection{Narrative-Specific Metrics}
The narrative specific metrics reveal a more nuanced picture of algorithm behavior in the context of topic discovery for intelligence applications. Narrative Distinctness measures the uniqueness and separability of discovered narrative themes. While the baseline achieves the highest score (0.352), DenStream (0.319) outperforms DBSTREAM (0.266), indicating that DenStream is more effective than DBSTREAM at identifying distinct topical narratives, though it does not surpass the baseline. The baseline HDBSCAN achieves the lowest Contingency (0.299) and Variance (0.286), suggesting stable, predictable cluster structures within each daily batch. In contrast, both online algorithms exhibit substantially higher values (DenStream: 0.389, 0.319; DBSTREAM: 0.630, 0.459), indicating greater variability in narrative structure across time. This increased contingency likely reflects the incremental nature of online algorithms, which adapt cluster structures as new documents arrive rather than performing global optimization on complete daily batches.

HDBSCAN produces three times more clusters than DenStream, even though the Number of Narrative Candidates differs between them by only a factor of two. This may indicate excessive granularity in HDBSCAN, manifested in the fragmentation of a single narrative into multiple distinct clusters. Conversely, DenStream may tend to merge several distinct narratives into a single cluster.

\subsection{Human Evaluation}
\begin{table}[t]
\centering
\small
\begin{tabular}{lcccc}
\hline
\textbf{Method} &
$\overline{\kappa}$ &
$\alpha$ &
\textbf{Accepted} &
\textbf{Accuracy} \\
\hline
HDBSCAN   & 0.731 & 0.733 & 183 / 236 & 0.775 \\
DBStream  & 0.784 & 0.781 &  94 / 116 & 0.810 \\
DenStream & \textbf{0.827} & \textbf{0.829} &  89 / 106 & \textbf{0.840} \\
\hline
\end{tabular}
\caption{
Human evaluation of narrative clustering quality.
Three annotators judged whether each generated narrative cluster was correct (true) or incorrect (false).
We report mean pairwise Cohen’s $\kappa$ ($\overline{\kappa}$), Krippendorff’s $\alpha$ (nominal),
and consensus-based acceptance using majority vote, and accuracy of the clustering based on acceptance rate.
}
\label{tab:narrative_clustering_eval}
\end{table}
Table~\ref{tab:narrative_clustering_eval} summarizes the human evaluation of narrative clustering quality across three clustering methods. Inter-annotator agreement is consistently high, with mean pairwise Cohen’s $\overline{\kappa}$ ranging from 0.73 to 0.83 and Krippendorff’s $\alpha$ between 0.73 and 0.83, indicating substantial to strong agreement.
Using majority vote as a proxy for cluster acceptance, DenStream achieves the highest accuracy (0.84), followed by DBStream (0.81) and HDBSCAN (0.78). These results indicate that DenStream produces narrative clusters that are more frequently judged as coherent and correct by human evaluators. Notably, methods with higher consensus acceptance rates also exhibit higher inter-annotator agreement. In particular, DenStream achieves both the highest acceptance rate and the strongest agreement($\overline{\kappa}=0.83$, $\alpha=0.83$), suggesting that its narrative clusters are not only more frequently judged as correct, but also less ambiguous and more consistently interpretable by human evaluators. %Overall, these results support the effectiveness of density-based streaming clustering for narrative induction, with DenStream producing the most stable and human-interpretable clusters in our setting.

\section{Discussion and Conclusion}
Overall, our results indicate that DenStream provides the most robust clustering performance, achieving clear improvements over the HDBSCAN baseline across both automatic clustering metrics and human evaluation. Its higher Silhouette Score and lower Davies–Bouldin Index suggest well-separated and cohesive clusters, which is further reflected in its superior Narrative Distinctness, highlighting its ability to recover meaningful and distinct thematic structures. Crucially, these structural advantages are corroborated by human evaluation: DenStream achieves the highest consensus acceptance rate (0.84) and the strongest inter-annotator agreement, indicating that its narrative clusters are not only quantitatively well-formed but also consistently interpretable and semantically coherent to human evaluators.
In contrast, DBStream performs poorly across standard clustering metrics, likely due to limitations of its micro-cluster representation in high-dimensional semantic embedding spaces. While DBStream achieves a moderate acceptance rate in human evaluation, its lower mean pairwise Cohen’s $\kappa$ compared to DenStream suggests that its cluster outputs are more ambiguous, making it harder for annotators to consistently agree on their correctness. This highlights an important distinction between acceptance frequency and interpretability: clusters may be judged acceptable in isolation while still lacking the clarity that enables stable human agreement.
The baseline HDBSCAN exhibits lower cluster distinctness in both automatic metrics and human acceptance rate, yet demonstrates greater temporal stability, as indicated by lower Narrative Contingency and Variance. This stability arises from complete daily re-clustering, which enforces consistent reference frames across time. However, human evaluation reveals that this stability does not necessarily translate into higher perceived cluster quality, suggesting that static re-clustering may preserve structural consistency at the expense of semantic sharpness.
Taken together, the experimental results reveal a fundamental tension between cluster quality, interpretability, and temporal stability. While DenStream introduces higher Narrative Contingency and Variance due to its incremental nature, human evaluation indicates that this variability corresponds to improved semantic clarity rather than noise. In dynamic intelligence settings, where evolving narratives are expected, such adaptability may be preferable to enforcing static topic definitions.
Finally, it is important to note that narrative-specific metrics may partially reflect biases introduced by the narrative extraction pipeline, which was fine-tuned on HDBSCAN outputs. As a result, some discrepancies observed for online algorithms may stem from a mismatch between clustering behavior and downstream expectations rather than intrinsic deficiencies in cluster quality. The stronger human validation of DenStream clusters suggests that these narrative metrics underestimate the semantic validity of online clustering approaches.

\section*{Limitations}
Several limitations should be considered when interpreting the results of this study.

All experiments were conducted on data from the Ukrainian information space, characterized by high geopolitical intensity, multilingual content (Russian, Ukrainian, code-switched \textit{surjik}), and elevated narrative volatility of wartime news stream. While this setting is representative of narrative monitoring applications, results may not directly generalize to domains with slower topic evolution, more formal language (e.g., scientific corpora), or lower noise levels. The relative advantages of online clustering are less pronounced in more static environments.

Clustering performance is inherently conditioned on upstream representation learning and dimensionality reduction. The experiments rely on transformer-based embeddings and UMAP projections; different encoder models, reduction techniques, or parameter settings could alter cluster geometry and affect the comparative performance of HDBSCAN and online methods. Thus, conclusions concern the clustering stage within this pipeline, not clustering in isolation.

Narrative-specific metrics are partly influenced by the LLM-based topic labeling and enrichment layer, which was originally tuned on HDBSCAN outputs. This may introduce structural bias favoring the baseline and partially penalizing online algorithms whose cluster shapes differ. Consequently, narrative metric comparisons may reflect both clustering behavior and downstream pipeline compatibility rather than pure cluster structure quality.

Some observed performance differences arise from practical limitations of available implementations (e.g., River), rather than intrinsic algorithmic properties. Although alternative implementations were used where necessary, system-level conclusions are influenced by software maturity and engineering trade-offs, which may change as libraries evolve.

The streaming simulation spans a limited historical window and a single evaluation day. While sufficient to study incremental behavior, longer-term dynamics such as gradual concept drift, seasonal topic cycles, or rare narrative shocks were not systematically evaluated. Extended longitudinal studies are required to fully assess stability and adaptation trade-offs.

Clusters are treated as proxies for narratives, yet narrative structure is inherently fluid and may not align perfectly with density-based cluster boundaries. Some differences in ``narrative quality” metrics may therefore reflect modeling assumptions rather than true differences in narrative discovery capability.

 DenStream's temporal variability might be mitigated through hyperparameter tuning or post-processing stabilization techniques, potentially enabling the algorithm to achieve both superior cluster quality and operational stability required for narrative monitoring applications. Additionally, retuning the narrative extraction pipeline to accommodate the characteristics of online clustering outputs may help reduce the observed narrative metric discrepancies and provide a more fair comparison across algorithms.
\bibliography{custom}

@article{mcinnes2017hdbscan,
  title   = {HDBSCAN: Hierarchical Density Based Clustering},
  author  = {McInnes, Leland and Healy, John and Astels, Steve},
  journal = {Journal of Open Source Software},
  volume  = {2},
  number  = {11},
  pages   = {205},
  year    = {2017},
  doi     = {10.21105/joss.00205}
}

@article{scarpino2022investigating,
  title   = {Investigating Topic Modeling Techniques to Extract Meaningful Insights in Italian Long COVID Narration},
  author  = {Scarpino, Irene and Zucco, Claudio and Vallelunga, Rosaria and Luzza, Francesco and Cannataro, Mario},
  journal = {BioTech},
  volume  = {11},
  number  = {3},
  pages   = {41},
  year    = {2022},
  doi     = {10.3390/biotech11030041}
}

@article{eklund2023empirical,
  title   = {An Empirical Configuration Study of a Common Document Clustering Pipeline},
  author  = {Eklund, Anton and Forsman, Mona and Drewes, Frank},
  journal = {Northern European Journal of Language Technology},
  volume  = {9},
  year    = {2023}
}

@incollection{chan1982updating,
  title     = {Updating Formulae and a Pairwise Algorithm for Computing Sample Variances},
  author    = {Chan, Tony F. and Golub, Gene H. and LeVeque, Randall J.},
  booktitle = {COMPSTAT 1982: Proceedings in Computational Statistics},
  editor    = {Caussinus, Henri and Ettinger, Philippe and Tomassone, Romain},
  publisher = {Physica-Verlag},
  address   = {Heidelberg},
  year      = {1982},
  doi       = {10.1007/978-3-642-51461-6_3}
}

@inproceedings{ester1996density,
  title     = {A Density-Based Algorithm for Discovering Clusters in Large Spatial Databases with Noise},
  author    = {Ester, Martin and Kriegel, Hans-Peter and Sander, J{\"o}rg and Xu, Xiaowei},
  booktitle = {Proceedings of the Second International Conference on Knowledge Discovery and Data Mining (KDD-96)},
  pages     = {226--231},
  publisher = {AAAI Press},
  year      = {1996}
}

@article{hahsler2016clustering,
  title   = {Clustering Data Streams Based on Shared Density Between Micro-Clusters},
  author  = {Hahsler, Michael and Bolanos, Matthew},
  journal = {IEEE Transactions on Knowledge and Data Engineering},
  volume  = {28},
  number  = {6},
  pages   = {1449--1461},
  year    = {2016},
  doi     = {10.1109/TKDE.2016.2522412}
}

@inproceedings{cao2006density,
  title     = {Density-Based Clustering over an Evolving Data Stream with Noise},
  author    = {Cao, Feng and Ester, Martin and Qian, Weining and Zhou, Aoying},
  booktitle = {Proceedings of the 2006 SIAM International Conference on Data Mining},
  pages     = {328--339},
  publisher = {SIAM},
  year      = {2006},
  doi       = {10.1137/1.9781611972764.29}
}

@article{das2025elda,
  title   = {eLDA: Augmenting Topic Modeling with Word Embeddings for Enhanced Coherence and Interpretability},
  author  = {Das, Panthadeep},
  journal = {Journal of Information Systems Engineering and Management},
  volume  = {10},
  pages   = {474--479},
  year    = {2025},
  doi     = {10.52783/jisem.v10i21s.3372}
}

@article{yadav2025hybrid,
  title   = {A Hybrid Model Integrating LDA, BERT, and Clustering for Enhanced Topic Modeling},
  author  = {Yadav, A. K. and Gupta, T. and Kumar, M. and others},
  journal = {Quality \& Quantity},
  volume  = {59},
  pages   = {2381--2408},
  year    = {2025},
  doi     = {10.1007/s11135-025-02077-y}
}

@article{angelov2020top2vec,
  title   = {Top2Vec: Distributed Representations of Topics},
  author  = {Angelov, Dimitar},
  journal = {arXiv preprint arXiv:2008.09470},
  year    = {2020}
}

@inproceedings{bianchi2021pretraining,
  title     = {Pre-training is a Hot Topic: Contextualized Document Embeddings Improve Topic Coherence},
  author    = {Bianchi, Federico and Terragni, Silvia and Hovy, Dirk},
  booktitle = {Proceedings of the 59th Annual Meeting of the Association for Computational Linguistics (ACL)},
  year      = {2021}
}

@incollection{caussinus1982computing,
  title     = {Computing Sample Variances},
  author    = {Caussinus, Henri and Ettinger, Philippe and Tomassone, Romain},
  booktitle = {COMPSTAT 1982},
  editor    = {Caussinus, H. and Ettinger, P. and Tomassone, R.},
  publisher = {Physica},
  address   = {Heidelberg},
  year      = {1982},
  doi       = {10.1007/978-3-642-51461-6_3}
}

@inproceedings{irani2025discourse,
  title     = {A Discourse Analysis Framework for Legislative and Social Media Debates},
  author    = {Irani, Arman and Park, Ju Yeon and Esterling, Kevin and Faloutsos, Michalis},
  booktitle = {Proceedings of the 17th ACM Web Science Conference (WebSci '25)},
  pages     = {199--209},
  publisher = {Association for Computing Machinery},
  address   = {New York, NY, USA},
  year      = {2025},
  doi       = {10.1145/3717867.3717918}
}

@article{upravitelev2025exploring,
  title   = {Exploring Contingency: Retrieving Disinformation Narratives with Speculative Fiction Generation},
  author  = {Upravitelev, Max and Solopova, Veronika and Jakob, Charlott and Sahitaj, Premtim and M{\"o}ller, Sebastian and Schmitt, Vera},
  journal = {Preprint},
  year    = {2025}
}

@article{mansalis2018evaluation,
  title   = {An Evaluation of Data Stream Clustering Algorithms},
  author  = {Mansalis, Stratos and Ntoutsi, Eirini and Pelekis, Nikos and Theodoridis, Yannis},
  journal = {Statistical Analysis and Data Mining},
  volume  = {11},
  number  = {4},
  pages   = {167--187},
  year    = {2018},
  doi     = {10.1002/sam.11380}
}

\appendix

\section{Implementation Challenges}
\label{sec:appendix}
During the implementation and experimental evaluation of clustering algorithms, several technical limitations were identified in commonly used reference implementations, substantially affecting performance and scalability, which should be taken into consideration when choosing a method. For DenStream, the predict\_one method in the River ML library recomputes final clusters by executing a DBSCAN-like procedure over all micro-clusters on every prediction call. While this behavior is consistent with fully online streaming assumptions, it leads to prohibitively slow prediction when the model state is static. This limitation is acknowledged by the River maintainers, who note that DenStream.predict\_one performs excessive recomputation for certain use cases. In offline evaluation settings, this repeated reclustering is unnecessary. Additionally, inconsistencies were observed in River’s DenstreamMicroCluster implementation, particularly in the maintenance of running weighted statistics, which can result in incorrect micro-cluster centers, radii, and weights.

For CluStream, performance issues arise during training rather than prediction. When micro-cluster merging is required, the algorithm performs quadratic-time pairwise distance computations between all micro-cluster centers, resulting in $O(M^2)$ complexity per merge event \cite{mansalis2018evaluation}. Combined with additional overhead from periodic k-means reclustering, this leads to rapidly increasing training time as the stream progresses. Due to these inherent scalability limitations, CluStream was fully rejected from further consideration in this study.
As a result of the identified performance and correctness issues, the River implementations were deemed unsuitable for large-scale experiments. Consequently, all DenStream-related results reported in this paper were obtained using pyDenStream, which more closely follows the original algorithmic specification and avoids repeated reclustering during prediction.

%This is an appendix.

\end{document}